\documentclass[runningheads]{llncs}
\PassOptionsToPackage{hyphens}{url}
\usepackage{graphicx}
\usepackage{xspace}

\usepackage{tikz}
\usepackage{comment}
\usepackage{amsmath,amssymb} 
\usepackage{color}
\usepackage{bm}
\usepackage{cite}
\usepackage{multirow}

\usepackage[accsupp]{axessibility}  

\usepackage[capitalize]{cleveref}
\crefname{section}{Sec.}{Secs.}
\Crefname{section}{Section}{Sections}
\Crefname{table}{Table}{Tables}
\crefname{table}{Tab.}{Tabs.}

\newcommand{\BSNlong}{BlindSpotNet\xspace}
\newcommand{\BSNshort}{BSN\xspace}
\newcommand{\BSSS}{\textit{Detection-2D}\xspace}
\newcommand{\BSBOX}{\textit{Detection-3D}\xspace}

\makeatletter
\DeclareRobustCommand\onedot{\futurelet\@let@token\@onedot}
\def\@onedot{\ifx\@let@token.\else.\null\fi\xspace}

\def\eg{\emph{e.g}\onedot} 
\def\ie{\emph{i.e}\onedot} 
\def\cf{\emph{cf}\onedot} 
\def\etc{\emph{etc}\onedot} 
\def\wrt{w.r.t\onedot} 
 
\def\etal{\emph{et al}\onedot}
\makeatother

\begin{document}
\pagestyle{headings}
\mainmatter
\def\ECCVSubNumber{3031}  

\title{\BSNlong: Seeing Where We Cannot See} 

\titlerunning{\BSNlong: Seeing Where We Cannot See}
%
\author{Taichi Fukuda \and
Kotaro Hasegawa \and 
Shinya Ishizaki \\
Shohei Nobuhara \and 
Ko Nishino}
\authorrunning{T. Fukuda et al.}
%
\institute{Graduate School of Informatics, Kyoto University}
\maketitle

\begin{abstract}
We introduce 2D blind spot estimation as a critical visual task for road scene understanding. By automatically detecting road regions that are occluded from the vehicle's vantage point, we can proactively alert a manual driver or a self-driving system to potential causes of accidents (\eg, draw attention to a road region from which a child may spring out). Detecting blind spots in full 3D would be challenging, as 3D reasoning on the fly even if the car is equipped with LiDAR would be prohibitively expensive and error prone. We instead propose to learn to estimate blind spots in 2D, just from a monocular camera. We achieve this in two steps. We first introduce an automatic method for generating ``ground-truth'' blind spot training data for arbitrary driving videos by leveraging monocular depth estimation, semantic segmentation, and SLAM. The key idea is to reason in 3D but from 2D images by defining blind spots as those road regions that are currently invisible but become visible in the near future. We construct a large-scale dataset with this automatic offline blind spot estimation, which we refer to as Road Blind Spot (RBS) dataset. Next, we introduce \BSNlong (\BSNshort), a simple network that fully leverages this dataset for fully automatic estimation of frame-wise blind spot probability maps for arbitrary driving videos. Extensive experimental results demonstrate the validity of our RBS Dataset and the effectiveness of our \BSNshort.

\end{abstract}


\section{Introduction}

Fully autonomous vehicles may soon become a reality. Advanced Driver-Assistance Systems (ADAS) have also become ubiquitous and intelligent. A large part of these developments is built on advancements in perceptual sensing, both in hardware and software. In particular, 3D and 2D visual sensing have played a large role in propelling these advancements. State-of-the-art LiDAR systems can resolve to centimeters at 300m or longer, and image understanding networks can recognize objects 100m away. A typical autonomous vehicle is equipped with up to ten of these sensors' visual perception, in addition to other modalities. 


These autonomous and assistive driving systems are, however, still prone to catastrophic errors, even when they are operating at low speeds~\cite{routers2021toyota}. More sensors would unlikely eliminate these errors. In fact, we human beings have much fewer visual sensors (just our two eyes) but can drive at least as well as current autonomous vehicles. Why are we able to do this? We believe, one of the primary reasons is that, although we are limited in our visual perception, we know that it's limited. That is, we are fully aware of when and where we can see and when and where we cannot see. We know that we can't see well at night so we drive cautiously; we know that we cannot see beyond a couple of hundred meters, so we don't drive too fast; we know that we cannot see behind us, so we use side and room mirrors. Most important, we know that we cannot see behind objects, \ie, that our vision can be obstructed by other objects in the scene. We are fully aware of these blind spots, so that we pay attention to those areas on the road with anticipation that something may spring out from them. That is, we know where to expect the unforeseen and we preemptively prepare for those events by attending our vision and mind to those blind spots. 


Can we make computers also ``see'' blind spots on the road? One approach would be to geometrically reason occlusions caused by the static and dynamic objects on the road (\eg, pedestrians and cars). This requires full 3D scene reconstruction and localization of the moving camera, which is computationally expensive and prone to errors as it requires fragile ray traversal. Instead, it would be desirable to directly estimate blind spots in the 2D images without explicit 3D reconstruction or sensing. This is particularly essential for driving safety, as we would want to detect blind spots on the fly as we drive down the road. Given an image (\ie, a frame from a live video), we would like to mark out all the road regions that are occluded by the camera, so that a driver or the autonomous system would know where to anticipate the unforeseen. 

How can we accomplish 2D blind spot detection? Just like we likely do, we could learn to estimate blind spots directly in 2D. The challenge for this lies in obtaining the training data--the ground truth. Again, explicit 3D reasoning is unrealistic as a perfect 3D reconstruction of every frame of a video would be prohibitively expensive and error prone, especially for a dynamic road scene in which blind spots change every frame. On the other hand, labeling blind spots in video frames is also near impossible, as even to the annotators, reasoning about the blind spots from 2D images is too hard a task, particularly for a dynamic scene. How can we then, formulate blind spot estimation as a learning task?

In this paper, we introduce a novel dataset and network for estimating occluded road regions in a driving video. We refer to the dataset as Road Blind Spot (RBS) Dataset and the network as \BSNlong (\BSNshort).
Our major contributions are two-fold: an algorithm for automatic generation of blind spot training data and a simple network that learns to detect blind spots from that training data. The first contribution is realized by implicitly reasoning 3D occlusions in a driving video from its 2D image frames by fully leveraging depth, localization, and semantic segmentation networks. This is made possible by our key idea of defining blind spots as areas on the road that are currently invisible but visible in the future. We refer to these as $T$-frame blind spots, \ie, those 2D road regions that are currently occluded by other objects that become visible $T$ frames later. Clearly, these form a subset of all true 2D blind spots; we cannot estimate the blind spots that never become seen in our video. They, however, cover a large portion of the blind spots (\cf \cref{fig:dataseteval}) that are critical in a road scene including those caused by parked cars on the side, oncoming cars on the other lane, street corners, pedestrians, and buildings. Most important, they allows us to derive an automatic means for computing blind spot regions for arbitrary driving videos. 

Our offline automatic training data generation algorithm computes $T$-frame blind spots for each frame of a driving video by playing it backwards, and by applying monocular depth estimation, SLAM, and semantic segmentation to obtain the depth, camera pose, semantic regions, and road regions. By computing the visible road regions in every frame, and then subtracting the current frame's from that of $T$-frames ahead, we can obtain blind spots for the current frame. A suitable value for $T$ can be determined based on the speed of the car and the frame rate of the video. Armed with this simple yet effective algorithm for computing blind spot maps, we construct the RBS Dataset. The dataset consists of 231 videos with blind spot maps computed for 21,662 frames.

For on-the-fly 2D blind spot estimation, we introduce \BSNlong, a deep neural network that estimates blind spot road regions for an arbitrary road scene directly for a single 2D input video frame, which fully leverages the newly introduced dataset. The network architecture is a fully convolutional encoder-decoder with Atrous Spatial Pyramid Pooling which takes in a road scene image as the input. We show that blind spot estimation can be implemented with a light-weight network by knowledge distillation from a semantic segmentation network. Through extensive experiments including ablation studies to analyze the effectiveness of the network architecture, we show that \BSNlong can accurately estimate the occluded road regions in any given frame independently, but resulting in consistent blind spot maps through a video. 

To the best of our knowledge, our work is the first to offer an extensive dataset and a baseline method for solving this important problem of 2D blind spot estimation. These results can directly be used to heighten the safety of autonomous driving and assisted driving, for instance, by drawing attention of the limited computation resource or the human driver to those blind spots (\eg, by sounding an alarm from the side with an oncoming large blind spot, if the person is looking away observed from an in-car camera). We also envision a future where \BSNlong helps in training better drivers both for autonomous and manual driving. The dataset and code will be made public once the paper is accepted to ensure timely progress towards such a future.

\section{Related Work}

Driver assistance around blind corners has been studied in the context of augmented/diminished reality~\cite{barnum2009dynamic,yasuda2012toward,rameau2016real}. Barnum \etal proposed a video-overlay system that presents a see-through rendering of hidden objects behind the wall for drivers using a surveillance camera installed at corners~\cite{barnum2009dynamic}. This system realizes realtime processing, but requires explicit modeling of blind spots beforehand. Our proposed method estimates them automatically.


Bird's-eye view (BEV) visualization~\cite{zhu2018generative,liu2020understanding} and scene parsing~\cite{Yang_2021_CVPR,sugiura2019probable,Mani_2020_WACV} around the vehicle can also assist drivers to avoid collision accidents.  Sugiura and Watanabe~\cite{sugiura2019probable} proposed a neural network that produces probable multi-hypothesis occupancy grid maps. Mani \etal~\cite{Mani_2020_WACV}, Yang \etal~\cite{Yang_2021_CVPR}, and Liu \etal~\cite{liu2020understanding} proposed road layout estimation networks. These methods, however, require sensitive 3D reasoning to obtain coarse blind spots at runtime, while our \BSNshort estimates blind spots directly in 2D without any 3D processing.

Amodal segmentation also handles occluded regions of each instance explicitly~\cite{zhu2017semantic,guo2012beyond,liu2016layered,song2017semantic,ehsani2018semantic,tighe2014scene,zhang2014panocontext}.  They segment the image into semantic regions while estimating occluded portions of each region. These methods, however, do not estimate objects that are completely invisible in the image.  We can find a side road while its road surface is not visible at all, \eg, due to cars parked in the street, by looking at gaps between buildings for example.

Understanding and predicting pedestrian behavior is also a primary objective of ADAS. Makansi \etal~\cite{makansi2020multimodal} trained a network that predicts pedestrians crossing in front of the vehicle.  Bertoni \etal~\cite{bertoni2019monoloco} estimated 3D locations of pedestrians around the subject by also modeling the uncertainty behind them. Our method explicitly recovers blind spots caused not only by pedestrians but also other obstacles including passing and parked cars, street medians, poles, etc.

\begin{figure*}[t]
    \centering
    \includegraphics[width=\linewidth]{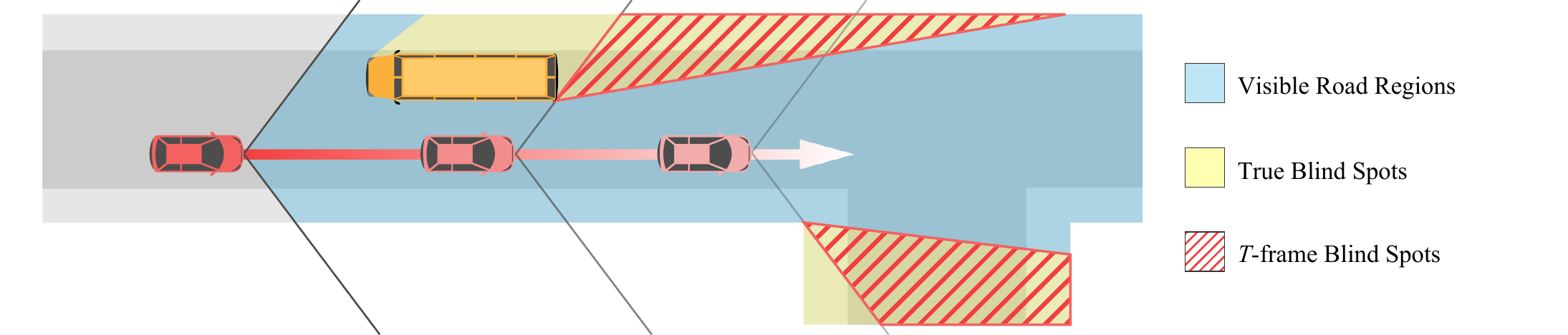}
    \caption{We define blind spots as road regions that are currently invisible but visible in the next $T$-frames, which we compute by aggregating traversable surface areas across future frames and negating the visible traversable surface area in the current frame. These $T$-frame blind spots form a subset of true blind spots, cover key regions of them and, most important, can directly be computed for arbitrary driving videos (\cf \cref{fig:dataseteval} for a real example). }
    \label{fig:topview}
\end{figure*}

Watson \etal~\cite{watson2020footprints} estimated free space including the area behind objects in the scene for robot navigation. They also generated a traversable area dataset to train their network. Our dataset generation leverages this footprint dataset generation algorithm to compute both visible and invisible road areas in a road scene image (\ie, driving video frame). Blind spot estimation requires more than just the area behind objects as it needs to be computed and estimated across frames with a dynamically changing viewpoint. 






\section{Road Blind Spot Dataset}
\label{sec:rbs}


\begin{figure*}[t]
\begin{center}
\includegraphics[width=1.0\linewidth]{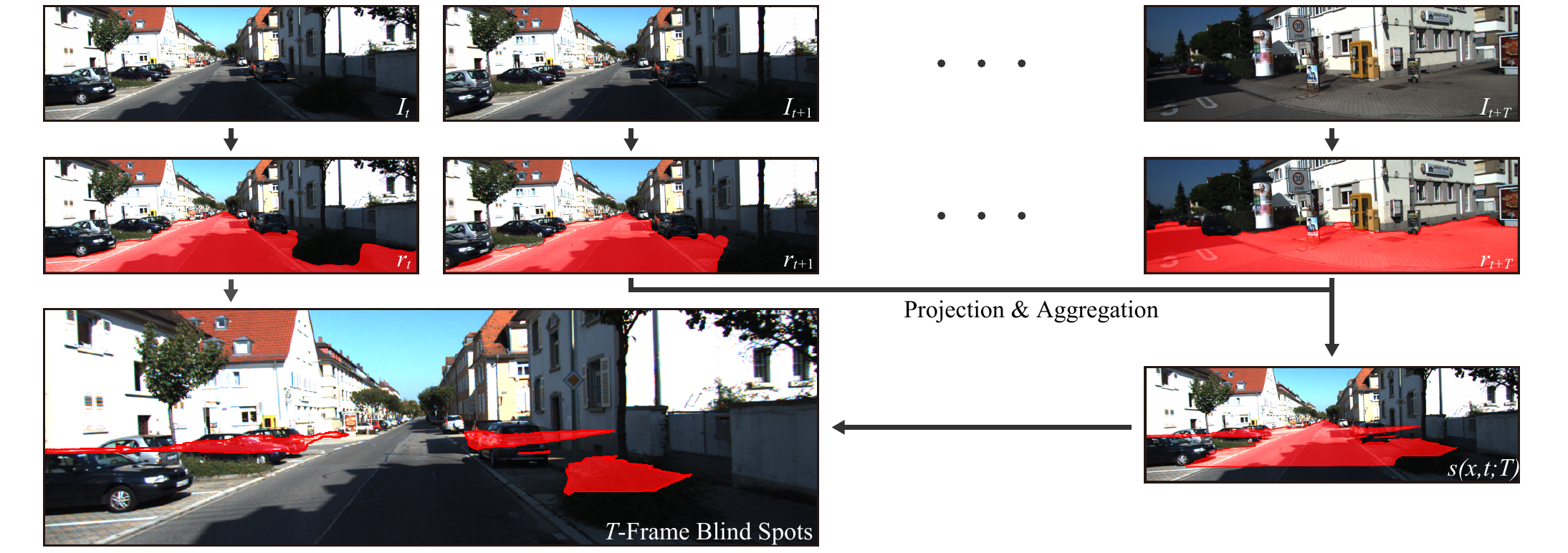}
\caption{Overview of algorithmic steps for generating $T$-frame blind spots for arbitrary driving video. Visible and invisible traversable surface maps are aggregated into the current frame from the next $T$ frames, from which the visible traversable region is negated to obtain the $T$-frame blind spots for the current frame.}
\label{fig:datageneration}
\end{center}
\end{figure*}
Our first goal is to establish a training dataset for learning to detect 2D blind spots. For this, we need an algorithm that can turn an arbitrary video into a video annotated with 2D blind spots for each and every frame. To derive such a method, we start by defining road-scene blind spots as something that we can compute from 2D videos offline and derive a method for computing those for arbitrary driving videos. With this algorithm, we construct a large-scale dataset of road blind spot videos (RBS Dataset). This data will later be used to train a network that can estimate 2D blind spots on the fly.

\subsection{$T$-Frame Blind Spots}

In the most general form, blind spots are volumes of the 3D space that are occluded from the viewpoint by an object in the scene. Computing these ``full blind volumes'' would be prohibitively expensive, especially for any application that requires real-time decision making. Even though our goal in this paper is not necessarily real-time computation at this point, a full 3D reasoning of occluded volumes would be undesirable as our target scenes are dynamic. 
We, instead, aim to estimate 2D blind spots on the road. Dangerous traffic situations are usually caused by unanticipated movements of dynamic objects (\eg, bikes, pedestrians, children, \etc) springing out from road areas invisible from the driver (or the camera of the car). 
Once we have the 2D blind spots, we can draw attention of the drivers by, for instance, extruding it perpendicularly to the road for 3D warning. By focusing on 2D blind spot estimation, we eliminate the need of explicit 3D geometric reasoning, which makes it particularly suitable for autonomous driving and ADAS applications.

As depicted in \cref{fig:topview}, given a frame of a driving video, we define our blind spots to be the road regions that are obstructed but becomes visible in the future. This clearly excludes blind spots that never become visible in any frame in the future, and thus the blind spots we compute and estimate are a subset of the true blind spots. That said, they have a good coverage and can be reliably computed from just ego-centric driving videos as we show in experimental results. Although we only investigate the estimation of these blind spots from ego-centric driving videos captured with regular perspective cameras, the wider the field of view, the better the coverage would become.

Formally, given a frame $I_t$, we define its blind spots as pixels $x\in \Omega_t^T$ corresponding to regions on the road $\Omega$ that are occluded but become visible in the next $T$ frames $\{I_{t+i} | i=1,\dots,T\}$. Our goal is to compute the set of pixels in the blind spots $\Omega_t$ as a binary mask of the image $\omega(x,t; T) : \mathbb{R}^2\times\mathbb{R} \mapsto \{0, 1\}$. Later \BSNlong will be trained to approximate this function $\omega(x,t;T)$.  We refer to the blind spots of this definition as \textit{$T$-frame blind spots}.

As depicted in \cref{fig:datageneration}, we compute the blind spot map $\omega$ by first computing and aggregating both visible and invisible traversable surface maps at frames $I_{t:t+T}$ and then eliminating the visible traversable surface map at target frame $I_t$. For this, similar to Watson \etal~\cite{watson2020footprints}, we compute a traversable surface map $s(x, t; T)$ by forward warping the traversable pixels from the next $T$ frames $\{I_{t+i} | i=1,\dots,T\}$ to the target frame $I_t$. To perform forward warping, we assume the camera intrinsics are known, the extrinsic parameter and the depth are estimated by a visual SLAM algorithm~\cite{sumikura2019openvslam} and an image-based depth estimation network~\cite{ranftl2020towards}, respectively.

Let $r(x,t)$ denote the visually-traversable region defined as a binary mask representing the union of the road and pavement areas estimated by a semantic segmentation as introduced in Watson \etal~\cite{watson2020footprints}.  We first project the visually-traversable regions estimated for $I_{t+i}$ ($i=1,\dots,T)$ to $I_t$ as $r'(x, t+i; t)$, and then aggregate them as
\begin{equation}
    s(x, t; T) = r'(x, t+1; t) \lor r'(x, t+2; t) \lor \dots \lor r'(x, t+T; t)\,, \label{eq:agg}
\end{equation}
where $\lor$ denotes the pixel-wise logical OR.
\begin{figure}[t]
    \centering
    \includegraphics[width=\linewidth]{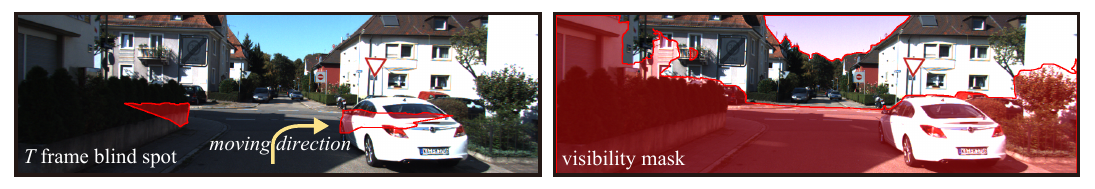}
    \caption{Visibility mask. Left: If the vehicle makes a turn, the blind spots straight ahead across the intersection never become visible. Right: To allow networks estimate such blind spots, we prepare a mask image to indicate the visible area for each frame.}
    \label{fig:mask}
\end{figure}
As blind spots are by definition invisible regions in frame $I_t$, the visually-traversable region $r(x,t)$ is deleted from $s(x, t; T)$ to obtain the final blind spots $\omega(x,t;T)$ by
\begin{equation}
    \omega(x,t; T) = s(x, t; T) \land \bar{r}(x, t)\,, \label{eq:omage_s_r}
\end{equation}
where $\land$ and $\bar{\cdot}$ denote the pixel-wise logical AND and negation, respectively.

Our method relies on three componential visual understanding tasks, namely semantic segmentation, monocular depth estimation, and visual SLAM. Although existing methods for these tasks achieve high accuracy, they can still suffer from slight errors. In the transformation from $r$ to $r'$, we use two such estimates, those of the camera pose and the depth, whose errors can cause residuals of visible road regions after \cref{eq:omage_s_r}.

We can rectify this with simple depth comparison. We first define aggregated depth mask $d_a$
\begin{equation}
d_a(x, t; T) = \frac{1}{M} \sum_{i = 1}^{T} r'(x, t+i; t) d'(x, t+i; t) \qquad
M = \sum_{i = 1}^{T} r'(x, t+i; t)\,,
\end{equation}
where $d'(x, t+i; t)$ is a depth map of frame $I_{t+i}$ projected onto frame $I_t$, which is calculated in a similar way as calculation of $r'(x, t+i; t)$. $M$ is the pixel-wise count of visually-traversable mask over $T$ frames.

When we compare the depth map $d(x)$ of frame $I_t$ with the aggregated depth mask $d_a$, the depth difference is large in the true blind spot region because true blind spots are occluded by foreground objects. On the other hand, it is small in erroneous blind spot regions because the compared depth values come from nearby pixels. Based on this observation, we remove erroneous blind spots by setting $\omega(x,t; T) = 0$ for the pixel $x$ that satisfy $|d(x) - d_a(x)| < l_d$. Here, $l_d$ is a threshold value determined empirically.

We may be left with small blind spots caused by, for instance, a shadow of a tire. These small blind spots are not important for driving safety. We remove these blind spots of less than 100 pixels from the final blind spot $\Omega_t$.


For building our RBS Dataset, we opt for MiDAS~\cite{ranftl2020towards} as the monocular depth estimator, OpenVSLAM~\cite{sumikura2019openvslam} for SLAM, and Panoptic-DeepLab~\cite{cheng2020panoptic} for semantic segmentation.  The scale of the depth estimated by MiDAS is linearly aligned with least squares fitting to the sparse 3D landmarks recovered by SLAM.


We use KITTI~\cite{Geiger2013IJRR}, BDD100k~\cite{yu2020bdd100k}, and TITAN~\cite{malla2020titan} datasets to build our RBS Dataset. By excluding videos for which the linear correlation coefficient in the MiDAS-to-SLAM depth alignment is less than 0.7, they provide 51, 62, and 118 videos, respectively.  We obtain blind spot masks for approximately 51, 34, and 12 minutes of videos in total, respectively.
The videos are resampled to 5 fps, and we set $T=5$ seconds for each video. We refer to them as KITTI-RBS, BDD100k-RBS, and TITAN-RBS Datasets, respectively.



\subsection{Visibility mask}
$T$-frame blind spots clearly do not cover blind spots that do not become visible through the video.  For example, as illustrated in \cref{fig:mask}, consider a frame where the vehicle is making a right turn.  The blind spots straight ahead across the intersection never become visible and hence are not included in the dataset, while \BSNlong should identify them once trained.  To disambiguate such invisible regions from non-blind spots, as shown in \cref{fig:mask}, we generate a binary mask $V_t$ called \textit{visibility mask} in addition to the blind spots $\Omega_t$.

We use semantic segmentation and the distance from the camera to define the visibility mask $V_t$.  For each pixel $x$, we first classify $x$ as visible, if the semantic segmentation label is ``sky.''  For non-sky pixels, we classify $x$ as visible if the minimum distance from the 3D point at distance $d(x,t)$ behind $x$ to the camera is less than a certain threshold $L$.  In our implementation, we set $L=16$ meters. 

\section{\BSNlong}

Now that we have (and can create limitless) abundant video data with per-frame blind spot annotations, we can formulate 2D blind spot detection as a learning problem. We derive a novel deep neural network for estimating blind spots in an image of a road scene. We refer to this network as \BSNlong and train and test it on our newly introduced RBS Dataset. 

\subsection{Network Architecture}

\begin{figure}[t]
    \centering
    \includegraphics[width=\linewidth]{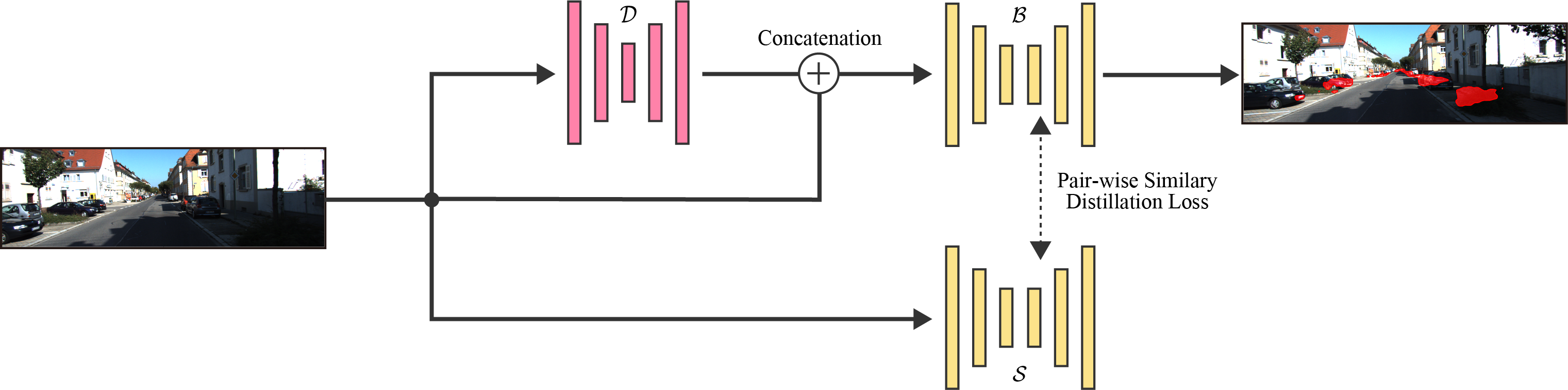}
    \caption{Overall architecture of \BSNlong. We leverage the fact that blind spot estimation bears similarity to semantic segmentation by adopting a light-weight network trained with knowledge distillation from a semantic segmentation teacher network.}
    \label{fig:network}
\end{figure}
As we saw in \cref{sec:rbs}, blind spots are mainly determined by the scene composition of objects including road regions and their ordering in 2D. As such, at a higher level, direct image-based estimation of blind spots shares similarity in its task to semantic segmentation. The task is, however, not necessarily easier, as it is 2D dense labeling but requires implicit 3D reasoning. Nevertheless, the output is a binary map (or its probability map), which suggests that a simpler network but with a similar inherent representation to semantic segmentation would be suitable for blind spot detection. 


\cref{fig:network} depicts the network architecture of \BSNlong. \BSNlong consists of three components: a depth estimator $\mathcal{D}$, a semantic segmentation teacher network $\mathcal{S}$, and a blind spot estimator $\mathcal{B}$.  Given an input image $I$ of size $W{\times}H{\times}3$, the depth estimator $\mathcal{D}$ estimates the depth map $D$ of size $W{\times}H$ from $I$.  The blind spot estimator takes both the RGB image $I$ and the estimated depth map $D$ as inputs and generates the blind spot map $B = \{b \in [0, 1] \}$.

The semantic segmentation network $\mathcal{S}$ serves as a teacher network to help train the blind spot estimator $\mathcal{B}$.  The blind spot estimator $\mathcal{B}$ should be trained to reason semantic information of the scene similar to semantic segmentation, but its output is abstracted as simple as a single-channel map $B$. This implies that training the blind spot estimator $\mathcal{B}$ only with the $T$-frame blind spots can easily bypass the semantic reasoning of the scene and overfit.  To mitigate this shortcut, we introduce the semantic segmentation network $\mathcal{S}$ pretrained on road scenes as a teacher and use its decoder output as a soft target of a corresponding layer output in the blind spot estimator $\mathcal{B}$.

\subsection{Knowledge Distillation}

Blind spot regions are highly correlated with the semantic structure of the scene. For instance, blind spots can appear behind vehicles and buildings, but never in visible road regions or in the sky. The blind spot estimator $\mathcal{B}$ should thus be able to learn useful representations from semantic segmentation networks for parsing road scenes when looking for blind spots. For this, we distill knowledge from a pretrained semantic segmentation network to our blind spot estimator $\mathcal{B}$.  Based on the work of Liu \etal~\cite{yifan2019structured}, we transfer the similarity between features at an intermediate layer of each network.

Suppose we subdivide the feature map of an intermediate layer of size $W' \times H' \times C$ into a set of $w' \times h'$ patches.  By denoting the spatial average of the features in the $i$th patch by $\bm{f}_i \in \mathbb{R}^C$, we define the similarity of two patches $i$ and $j$ by their cosine distance
    $a_{ij} = \frac{\bm{f}_i^{\top} \bm{f}_j}{\|\bm{f}_i\|  \|\bm{f}_j\|}.$
Following \cite{yifan2019structured}, given this pairwise similarity for patches in both the teacher semantic segmentation network $\mathcal{S}$ and the student network, \ie, the blind spot estimator $\mathcal{B}$, as $a_{ij}^\mathcal{S}$ and $a_{ij}^\mathcal{B}$, respectively, we introduce a pair-wise similarity distillation loss $l_\mathrm{KD}$ as
\begin{equation}
    l_\mathrm{KD} = \frac{1}{\left(w' \times h'\right)^2} \sum_{i \in \mathcal{R}} \sum_{j \in \mathcal{R}} \left(a_{ij}^{\mathcal{S}} - a_{ij}^{\mathcal{B}}\right)^2\,,
    \label{eq:loss_kd}
\end{equation}
where $\mathcal{R} = \{1, 2, \dots, w' \times h' \}$ denotes the entire set of patches.  In our implementation, we opted for DeepLabV3+~\cite{chen2018encoder} as the teacher network $\mathcal{S}$.

\subsection{Loss Function}
\label{sec:loss}

In addition to the similarity distillation loss $l_\mathrm{KD}$ in \cref{eq:loss_kd}, we employ a binary cross entropy loss $l_\mathrm{BCE}$ between the output of the blind spot estimator $\mathcal{B}$ and the $T$-frame blind spots given by our RBS Dataset as
\begin{equation}
    l_\mathrm{BCE} = - \frac{1}{|V|} \sum_{x \in V} (\omega(x) \log b(x) + (1 - \omega(x)) \log (1 - b(x)))\,,
\end{equation}
where $x$ denotes the pixels in the visibility mask $V$, $\omega(x)$ and $b(x)$ denote the $T$-frame blind spots and the estimated probabilities at pixel $x$. $|V|$ is the total number of the pixels in $V$.  The total loss function $L$ to be optimized is defined as a weighted sum of these two loss functions 
$L = l_\mathrm{BCE} + \lambda l_\mathrm{KD}$, 
where $\lambda$ is an empirically determined weighting factor.

\section{Experimental Results}

We evaluate the validity of RBS Dataset and the effectiveness of \BSNlong (\BSNshort), qualitatively and quantitatively with a comprehensive set of experiments.



\subsection{RBS Dataset Evaluation}

\begin{table}[t]
    \centering
    \caption{Quantitative evaluation of RBS Datasets.  The two numbers in each cell indicate the recall and the false-negative rate \wrt the ground truth blind spot. Please see \cref{fig:dataseteval} for visualization. The results show that our $T$-frame blind spots approximate true blind spots well.}
    \setlength{\tabcolsep}{4pt}
    \label{tab:rbs_table}
    \begin{tabular}{lcc|cc}
         & \multicolumn{2}{c|}{CARLA} & \multicolumn{2}{c}{KITTI} \\
         & Rec.$\uparrow$ & FN rate$\downarrow$ & Rec.$\uparrow$ & FN rate$\downarrow$ \\
         \hline \hline 
         $T$-frame BS-GT & 0.372 & 0.013 & \multicolumn{2}{c}{N/A\footnotemark} \\ 
         $T$-frame BS (ours) & 0.297 & 0.015 & 0.169 & 0.056 \\
    \end{tabular}
\end{table}

    \footnotetext[1]{$T$-frame BS-GT is not available since KITTI does not have ground truth semantic segmentation for most of the frames.}

\begin{figure}[t]
    \centering

        \includegraphics[width=\linewidth]{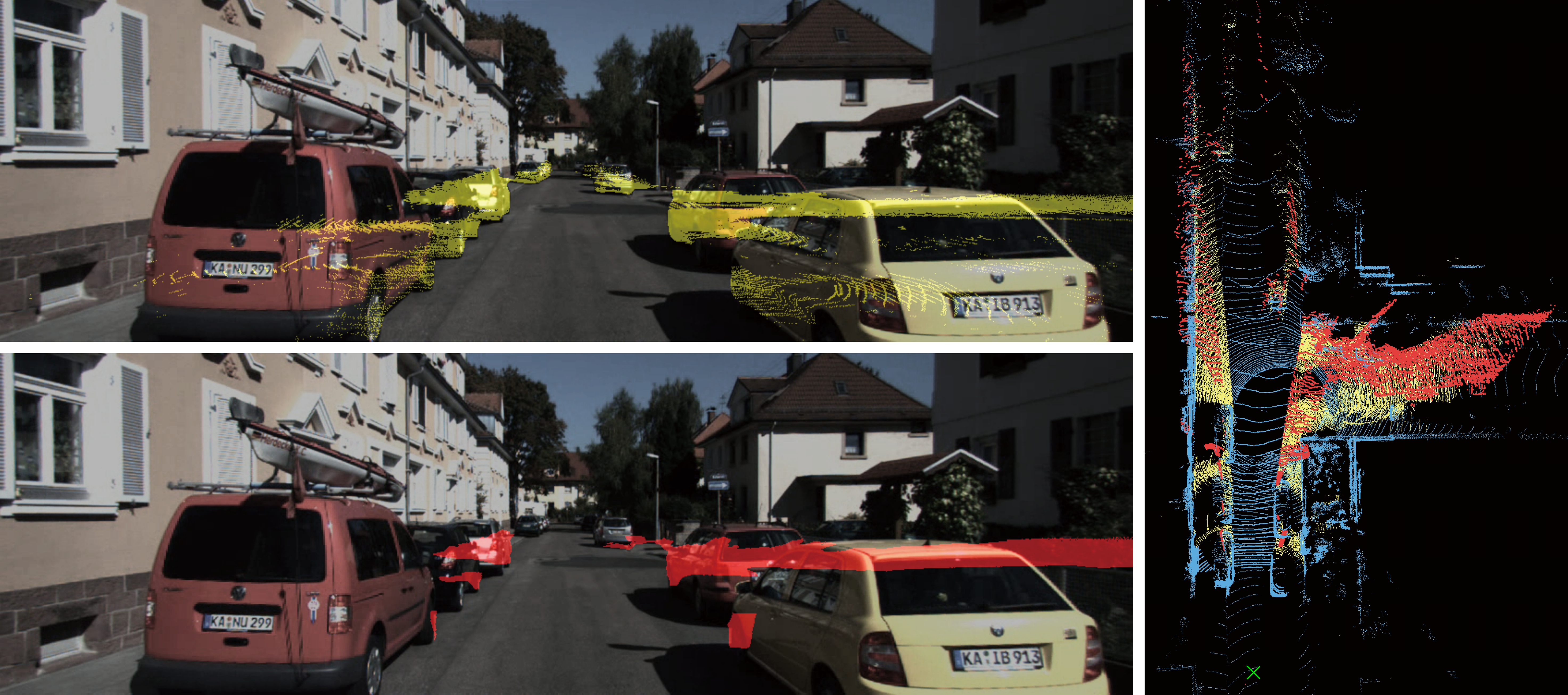}
    \caption{Comparison of $T$-frame blind spot and ``ground truth'' obtained from LiDAR data. Top-left and bottom-left figures show the ground truth and $T$-frame blind spots, respectively. Right figure shows the 3D projection of ground truth (yellow) and $T$-frame blind spots (red) to LiDAR scan (blue). The green cross mark indicates the camera / LiDAR position.}
    \label{fig:dataseteval}
\end{figure}

We first validate our $T$-frame blind spots with synthetic data generated by CARLA~\cite{dosovitskiy2017carla} and with real data from KITTI~\cite{Geiger2013IJRR}. How well do they capture true blind spots?
We use $360^\circ$ LiDAR scans to obtain ground-truth blind spots (BS-GT). Note again that for real use such ground truth computation will be prohibitively expensive and would require LiDAR. We also use the ground-truth depth maps to compute $T$-frame blind spots without noise ($T$-frame BS-GT). In computing BS-GT, we find road regions in LiDAR points by fitting the road plane manually in 3D. For $T$-frame BS-GT, we used the ground truth semantic labels.  We compare the $T$-frame blind spots generated by our data generation algorithm ($T$-frame BS)  and $T$-frame BS-GT with BS-GT, and evaluate its quality in terms of the recall and the false-negative rate. \cref{tab:rbs_table} and \cref{fig:dataseteval} show the results.  Since BS-GT is defined by sparse LiDAR points while $T$-frame BS-GT and $T$-frame BS use dense depth-maps, the precision and the false-positive (type-I error) rate do not make sense. These results show that our $T$-frame blind spots approximates the ground-truth blind spot well.

\subsection{\BSNlong Evaluation}


\paragraph{\BSNlong}

We use MiDAS~\cite{ranftl2020towards} and DeepLabV3+~\cite{chen2018encoder} as the depth estimator $\mathcal{D}$ and the semantic segmentation subnetwork $\mathcal{S}$, respectively. We use ResNet-18 and Atrous Spatial Pyramid Pooling~\cite{chen2018encoder} for the blind spot estimator $\mathcal{B}$ following DeepLabV3+. \BSNlong is trained by Adam~\cite{kingma2014adam} with $\beta_1=0.9, \beta_2=0.999, \epsilon=1 \times 10^{-8}$, and weight decay $5\times 10^{-4}$. The learning rate is initialized to $0.001$ and a polynomial scheduler is applied. The knowledge distillation coefficient $\lambda$ in \cref{sec:loss} is fixed to $1.0$. 

We divided the 10,135 frames from KITTI-RBS into training, validation, and test sets by $8:1:1$, and the 8,872 frames from BDD100k-RBS into training and test sets by $8:2$.  We used all the 2,655 frames from TITAN-RBS for evaluation only since each video is too short (10 to 20 seconds) to be used for training.


\vspace{-12pt}
\paragraph{Metrics}

Blind spot estimation is a binary segmentation problem.  For each input frame, our \BSNlong outputs the probability map of blind spots.  We threshold this probability map to obtain the final binary blind spot mask, and compare it with the $T$-frame blind spots by IoU, recall, and precision. The threshold is determined empirically for each dataset as it depends on the road scene. We plan to learn this threshold as part of the network in future work.
Notice that our RBS Datasets include blind spots that becomes visible in $T$ frames only.  For IoU, recall, and precision, we only consider pixels in the visibility mask. 


\vspace{-12pt}
\paragraph{Baselines}
As discussed earlier, our work is the first for image-based 2D blind spot estimation, and there are no other existing methods in the literature to the best of our knowledge.  For comparison, we adapt state-of-the-art traversable region estimation~\cite{watson2020footprints}, 2D vehicle/pedestrian/cyclist detection by semantic segmentation~\cite{chen2018encoder}, and 3D vehicle/pedestrian/cyclist detection~\cite{brazil2019m3drpn} for 2D blind spot estimation as baselines and refer to them as \textit{Traversable}, \BSSS, and \BSBOX, respectively.

For \textit{Traversable} we use the hidden traversable regions, estimated by the original implementation of Watson \etal~\cite{watson2020footprints}, as blind spots. \BSSS is a simple baseline that detects vehicle, pedestrian, and cyclist regions estimated by DeepLabV3+~\cite{chen2018encoder} as blind spots. \BSBOX utilizes a single-image 3D detection of vehicles, pedestrians, and cyclists by Brazil \etal~\cite{brazil2019m3drpn}. Given their detected 3D bounding boxes, \BSBOX returns the intersection of the projection of their far-side faces and the results by \BSSS as blind spots.

We also compare with \BSNshort without depth estimation (\BSNshort-D), and \BSNshort without knowledge distillation (\BSNshort-KD) for ablation studies.  In \BSNshort-D, the depth estimator $\mathcal{D}$ is removed from \BSNshort, and the blind spot estimator $\mathcal{B}$ is modified to take the original RGB image directly.  \BSNshort-KD disables the knowledge distillation loss by setting $\lambda=0$ in \cref{sec:loss} in \BSNshort.

\subsubsection{Quantitative Evaluations}

\begin{table}[t]
    \centering
    \caption{Quantitative results on the test sets from KITTI-RBS, BDD100k-RBS, and TITAN-RBS.  The lines with ``w/ KITTI-RBS'' and ``w/ BDD100k-RBS'' report the results by the networks trained with KITTI-RBS and BDD100k-RBS Datasets, respectively.  These results show that depth estimation and knowledge distillation contribute independently to the final accuracy, and our \BSNshort successfully estimates blind spots across different scenes (\ie, datasets).}
    \label{tab:quantitative}
    \begingroup
    \setlength{\tabcolsep}{4pt}
    \tiny
    \begin{tabular}{lccc|ccc|ccc}
        & \multicolumn{3}{c}{KITTI-RBS} & \multicolumn{3}{|c|}{BDD100k-RBS} & \multicolumn{3}{c}{TITAN-RBS} \\
        Model & IoU$\uparrow$ & Rec.$\uparrow$ & Prec.$\uparrow$ & IoU$\uparrow$ & Rec.$\uparrow$ & Prec.$\uparrow$ & IoU$\uparrow$ & Rec.$\uparrow$ & Prec.$\uparrow$ \\
        \hline\hline
        \textit{Traversable} based on ~\cite{watson2020footprints} & 0.176 & 0.462 & 0.222 & 0.088 & 0.198 & 0.136 & 0.135 & 0.303 & 0.196 \\
        \BSSS based on ~\cite{cheng2020panoptic} & 0.129 & 0.581 & 0.142 & 0.184 & \textbf{0.652} & 0.204 & 0.142 & 0.435 & 0.182 \\
        \BSBOX based on ~\cite{brazil2019m3drpn} & 0.182 & 0.368 & 0.265 & 0.059 & 0.067 & 0.316 & 0.048 & 0.057 & 0.216 \\
        \hline
        \BSNshort-D w/ KITTI-RBS & 0.296 & 0.601 & 0.368 & 0.295 & 0.438 & 0.475 & 0.250 & 0.474 & 0.345 \\
        \BSNshort-KD w/ KITTI-RBS & 0.305 & \textbf{0.646} & 0.367 & 0.225 & 0.249 & \textbf{0.700} & 0.168 & 0.249 & 0.342 \\
        \BSNshort (Ours) w/ KITTI-RBS & \textbf{0.330} & 0.563 & 0.444 & 0.283 & 0.349 & 0.599 & 0.187 & 0.280 & 0.360 \\
        \hline
        \BSNshort-D w/ BDD100k-RBS & 0.270 & 0.629 & 0.321 & 0.360 & 0.478 & 0.593 & 0.244 & 0.420 & 0.367 \\
        \BSNshort-KD w/ BDD100k-RBS & 0.187 & 0.210 & \textbf{0.633} & 0.350 & 0.443 & 0.624 & 0.253 & 0.529 & \textbf{0.326} \\
        \BSNshort (Ours) w/ BDD100k-RBS & 0.314 & 0.599 & 0.398 & \textbf{0.364} & 0.533 & 0.535 & \textbf{0.257} & \textbf{0.554} & 0.324 \\
    \end{tabular}
    \endgroup
\end{table}

\cref{tab:quantitative} shows the results on the test sets from KITTI-RBS, BDD100k-RBS, and TITAN-RBS Datasets.  The lines with ``w/ KITTI-RBS'' and ``w/ BDD100k-RBS'' indicate the results of the networks trained with KITTI-RBS and BDD100k-RBS Datasets, respectively. Each network, after pre-training, was fine-tuned using 20\% of the training data of the target dataset to absorb strong scene biases. It is worth mentioning that this fine-tuning is closer to self-supervision as the $T$-frame blind spots can be automatically computed without any external supervision for arbitrary videos. As such, \BSNlong can be applied to any driving video without suffering from domain gaps, as long as a small amount of video can be acquired before running \BSNlong for inference. The 20\% training data usage of the target scene simulates such a scenario. Note that none of the test data were used and this fine-tuning was not done for TITAN-RBS.

\BSNshort outperforms Watson \etal~\cite{watson2020footprints}, \BSSS, and \BSBOX.  These results show that blind spot estimation cannot be achieved by simply estimating footprint or ``behind-the-vehicle/pedestrian/cyclist'' regions.
The full \BSNshort also performs better than \BSNshort-D and \BSNshort-KD.  This suggests that both the depth estimator and knowledge distillation contribute to its performance independently. Furthermore, the performance of \BSNshort w/ KITTI-RBS on BDD100k-RBS and TITAN-RBS and that of \BSNshort w/ BDD100k-RBS on KITTI-RBS and TITAN-RBS demonstrate the ability of \BSNshort to generalize across datasets.

{\arrayrulewidth=1pt
\begin{figure}[t]
    \centering
    \begin{tikzpicture}[x=0.001\linewidth,y=0.001\textwidth]
        \newcount\vX
        \newcount\vY
        \newcount\vdX
        \newcount\vdY
        \def\INC#1{\includegraphics[width=0.31\linewidth]{#1}}
        \vdX = 315
        \vdY = -105
        \vX = 0
        \vY = 0
        \node[inner sep=0pt, rotate=90,font=\tiny] (a) at (\vX,\vY) {\parbox{1.2cm}{$T$-Frame \\[-1pt] blind spots}\rule{0pt}{10pt}};
        \vX = 175
        \node[inner sep=0pt] (a) at (\vX, \vY) {\INC{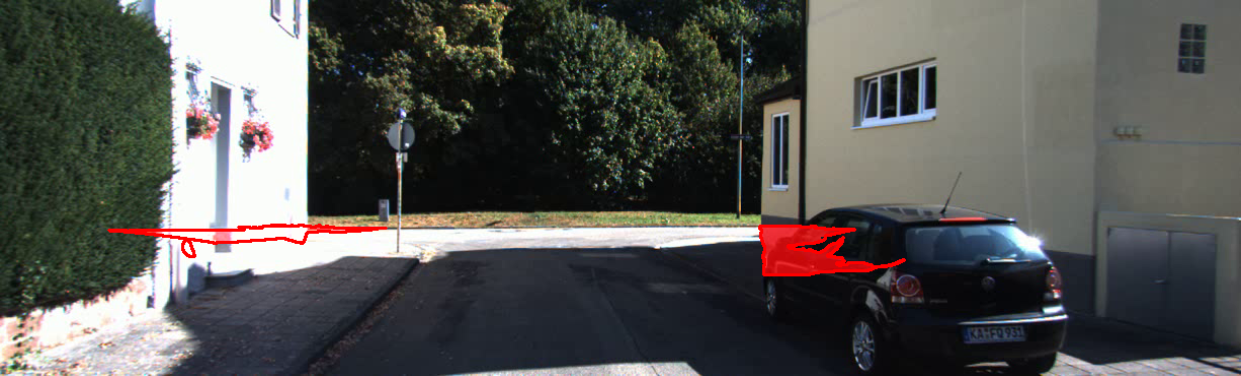}};
        \advance\vX by \vdX
        \node[inner sep=0pt] (a) at (\vX, \vY) {\INC{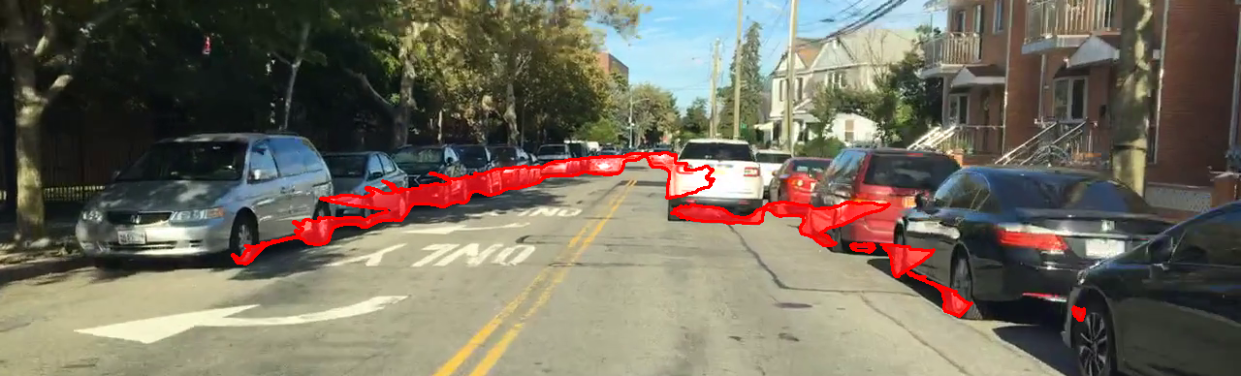}};
        \advance\vX by \vdX
        \node[inner sep=0pt] (a) at (\vX, \vY) {\INC{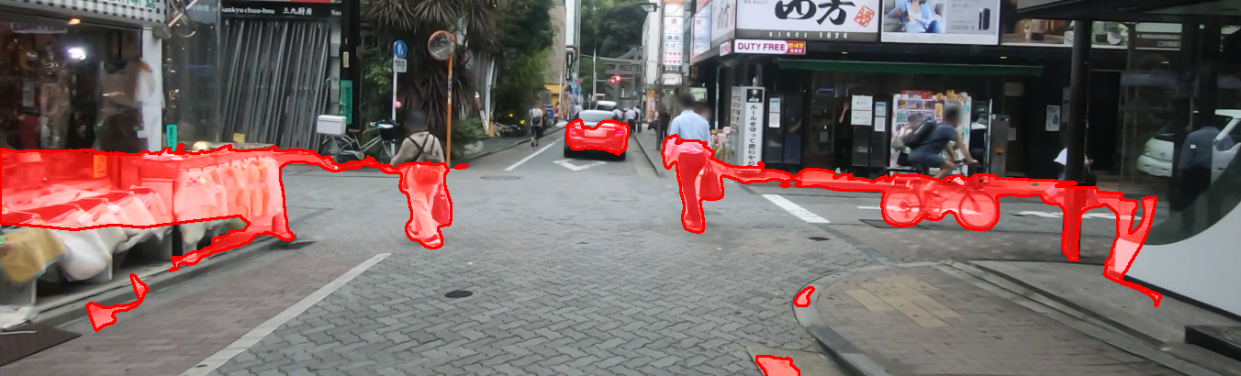}};

        \vX = 0
        \advance\vY by \vdY
        \node[inner sep=0pt, rotate=90,font=\tiny] (a) at (\vX,\vY) {\textit{Traversable}\rule{0pt}{12pt}};
        \vX = 175
        \node[inner sep=0pt] (a) at (\vX, \vY) {\INC{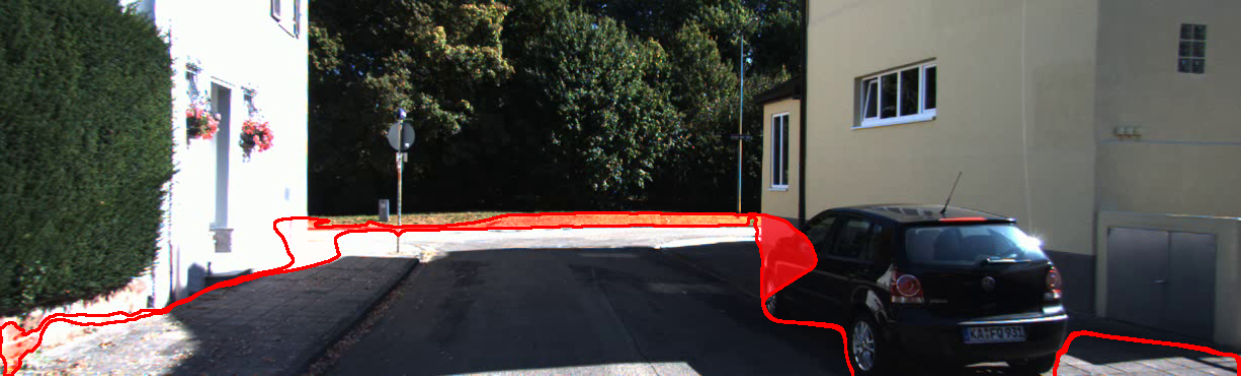}};
        \advance\vX by \vdX
        \node[inner sep=0pt] (a) at (\vX, \vY) {\INC{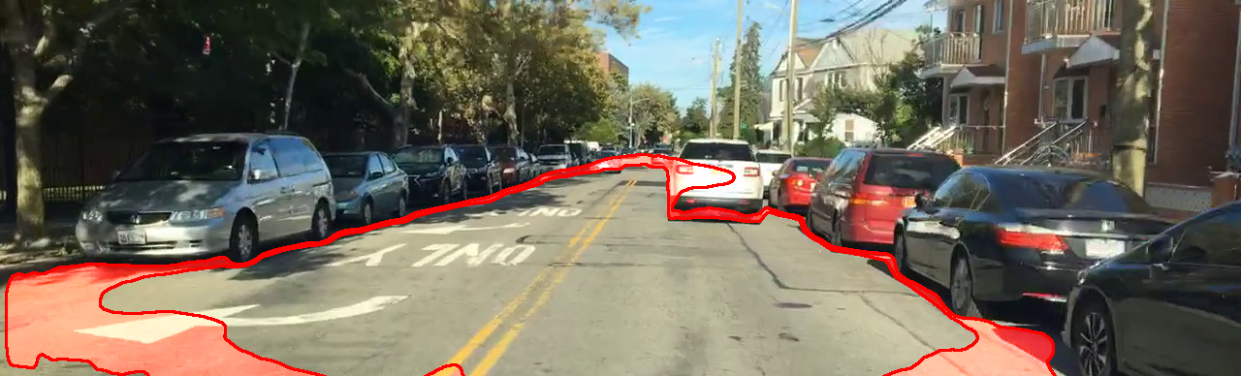}};
        \advance\vX by \vdX
        \node[inner sep=0pt] (a) at (\vX, \vY) {\INC{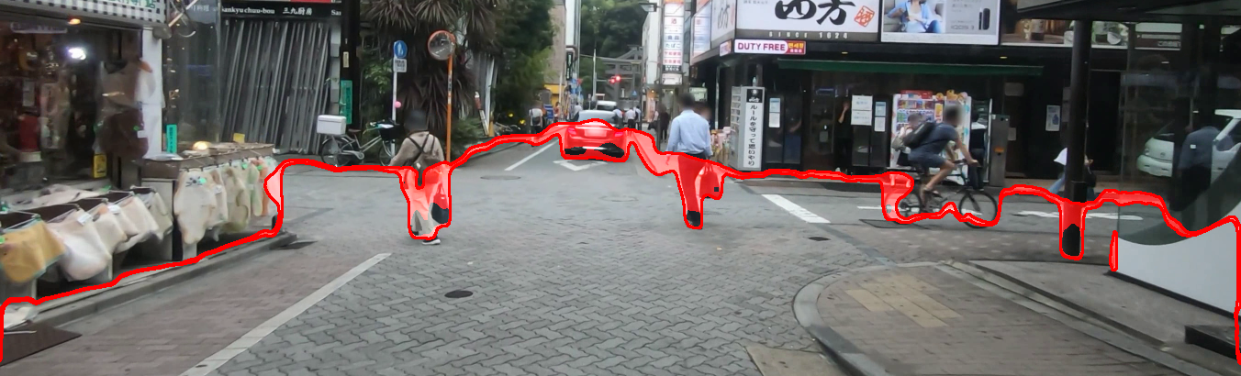}};

        \vX = 0
        \advance\vY by \vdY
        \node[inner sep=0pt, rotate=90,font=\tiny] (a) at (\vX,\vY) {\BSSS\rule{0pt}{12pt}};
        \vX = 175
        \node[inner sep=0pt] (a) at (\vX, \vY) {\INC{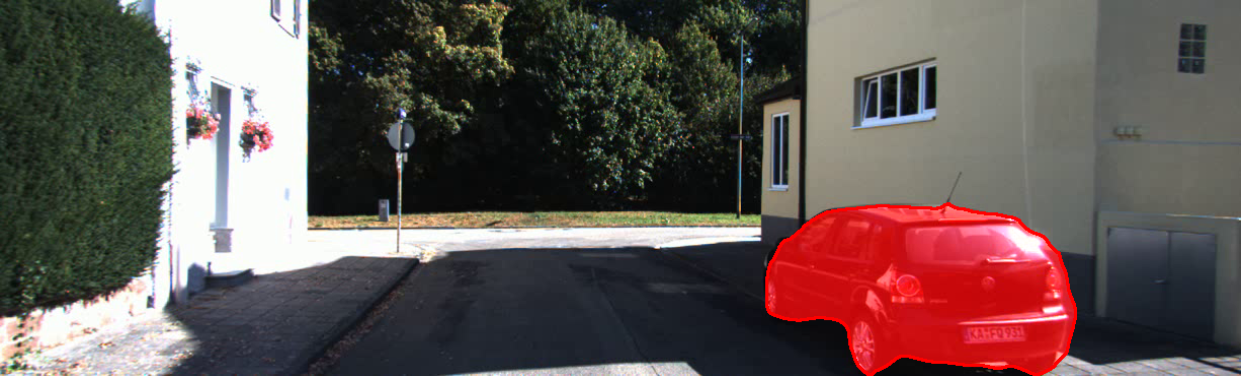}};
        \advance\vX by \vdX
        \node[inner sep=0pt] (a) at (\vX, \vY) {\INC{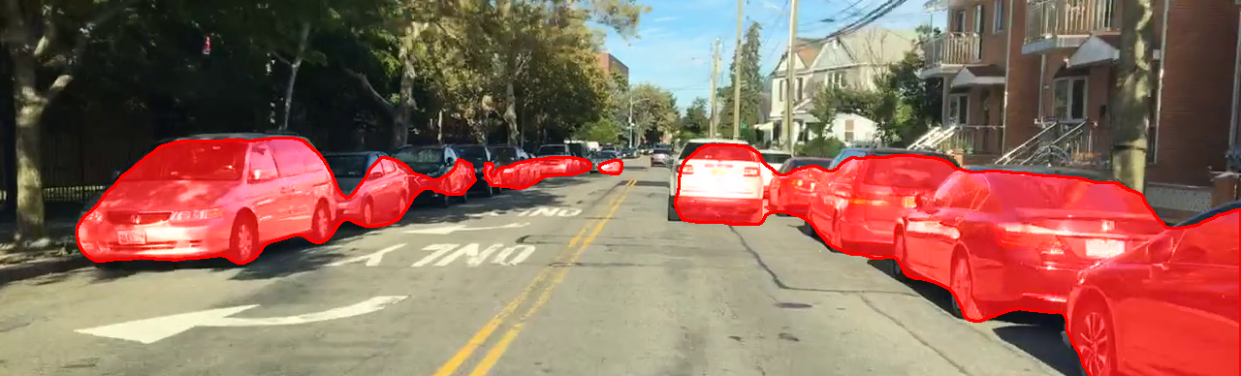}};
        \advance\vX by \vdX
        \node[inner sep=0pt] (a) at (\vX, \vY) {\INC{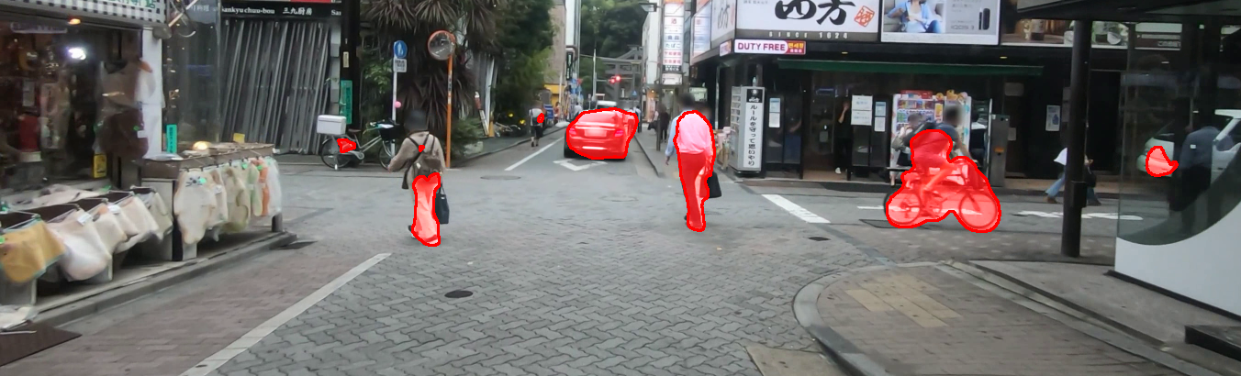}};

        \vX = 0
        \advance\vY by \vdY
        \node[inner sep=0pt, rotate=90,font=\tiny] (a) at (\vX,\vY) {\BSBOX\rule{0pt}{12pt}};
        \vX = 175
        \node[inner sep=0pt] (a) at (\vX, \vY) {\INC{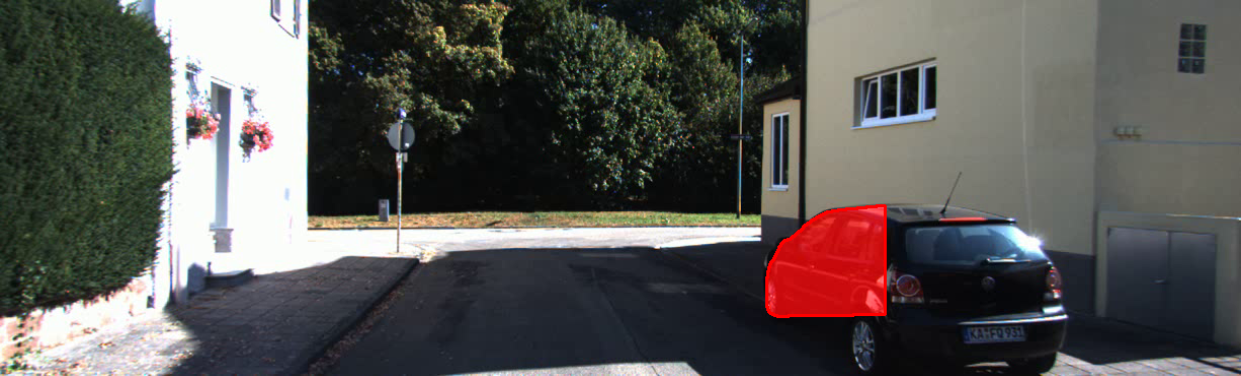}};
        \advance\vX by \vdX
        \node[inner sep=0pt] (a) at (\vX, \vY) {\INC{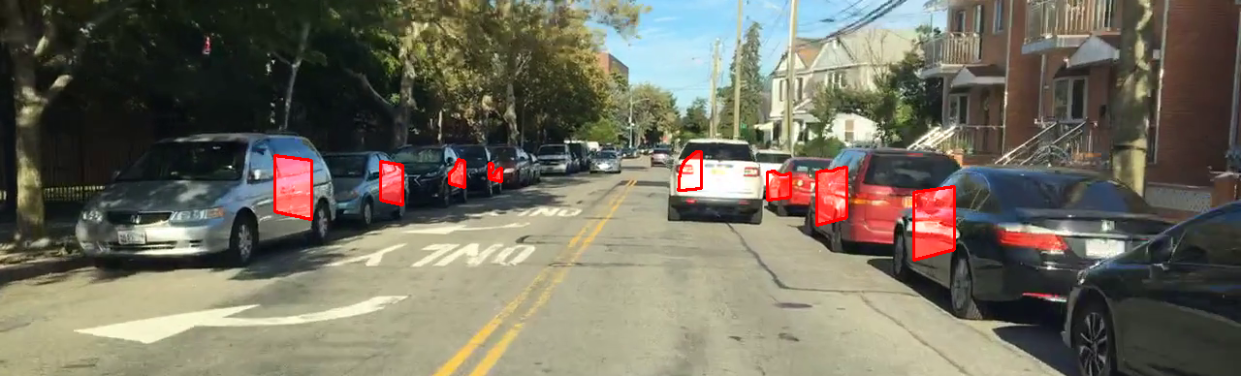}};
        \advance\vX by \vdX
        \node[inner sep=0pt] (a) at (\vX, \vY) {\INC{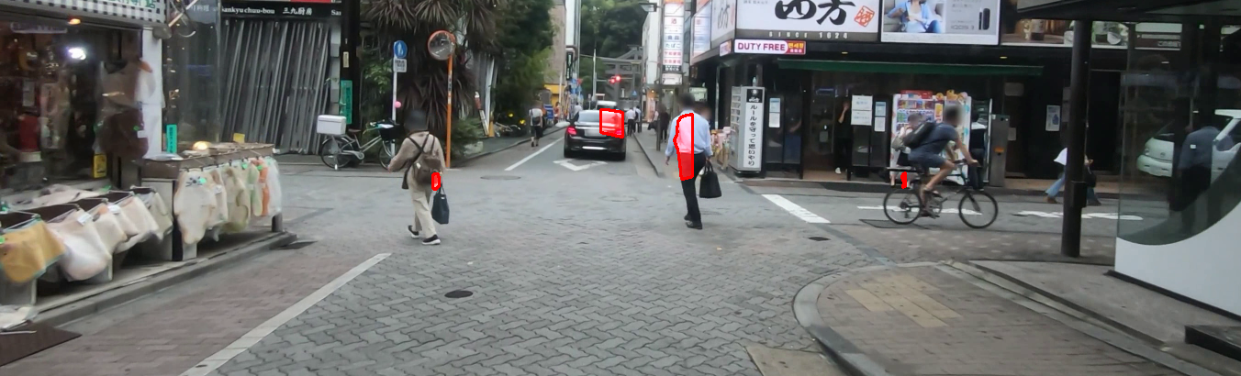}};

        \vX = 0
        \advance\vY by \vdY
        \node[inner sep=0pt, rotate=90,font=\tiny] (a) at (\vX,\vY) {\BSNshort-D\rule{0pt}{12pt}};
        \vX = 175
        \node[inner sep=0pt] (a) at (\vX, \vY) {\INC{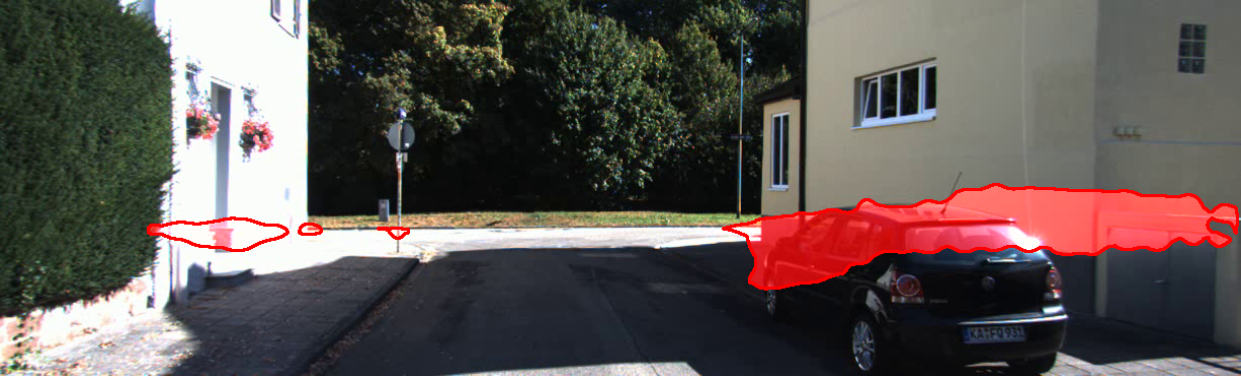}};
        \advance\vX by \vdX
        \node[inner sep=0pt] (a) at (\vX, \vY) {\INC{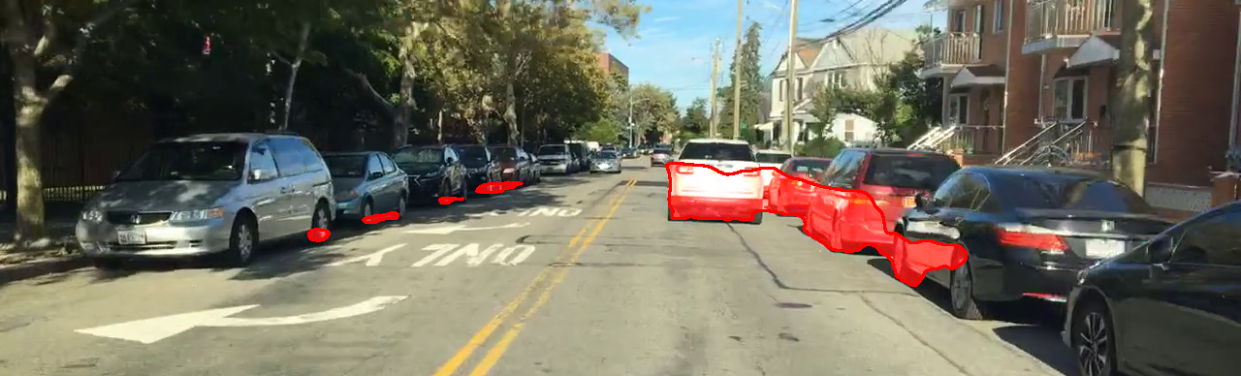}};
        \advance\vX by \vdX
        \node[inner sep=0pt] (a) at (\vX, \vY) {\INC{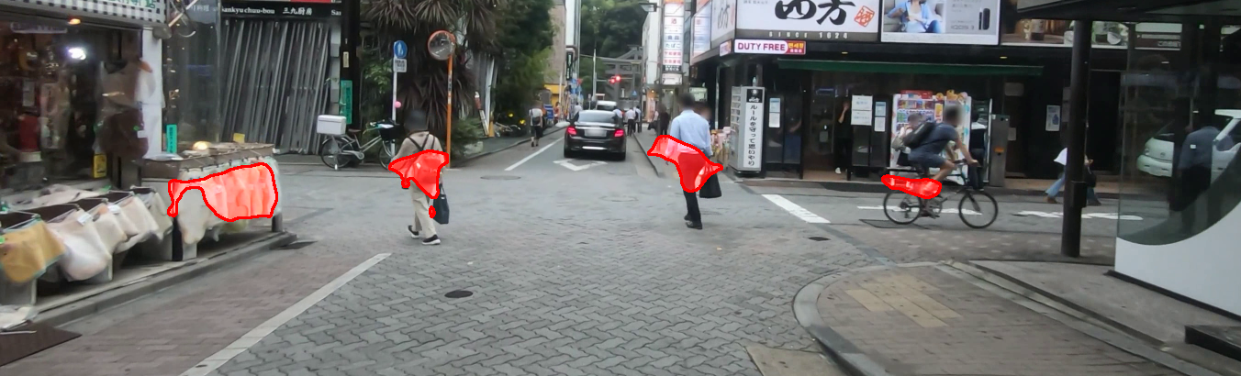}};

        \vX = 0
        \advance\vY by \vdY
        \node[inner sep=0pt, rotate=90,font=\tiny] (a) at (\vX,\vY) {\BSNshort-KD\rule{0pt}{12pt}};
        \vX = 175
        \node[inner sep=0pt] (a) at (\vX, \vY) {\INC{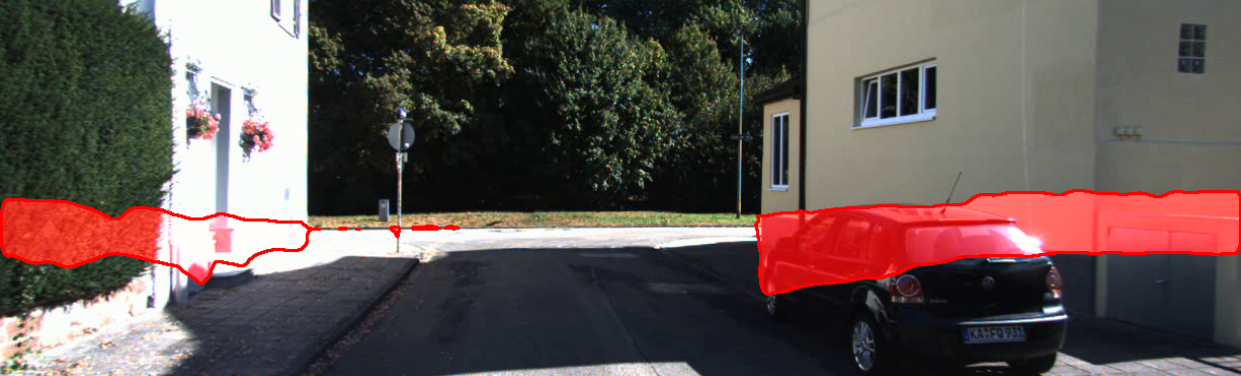}};
        \advance\vX by \vdX
        \node[inner sep=0pt] (a) at (\vX, \vY) {\INC{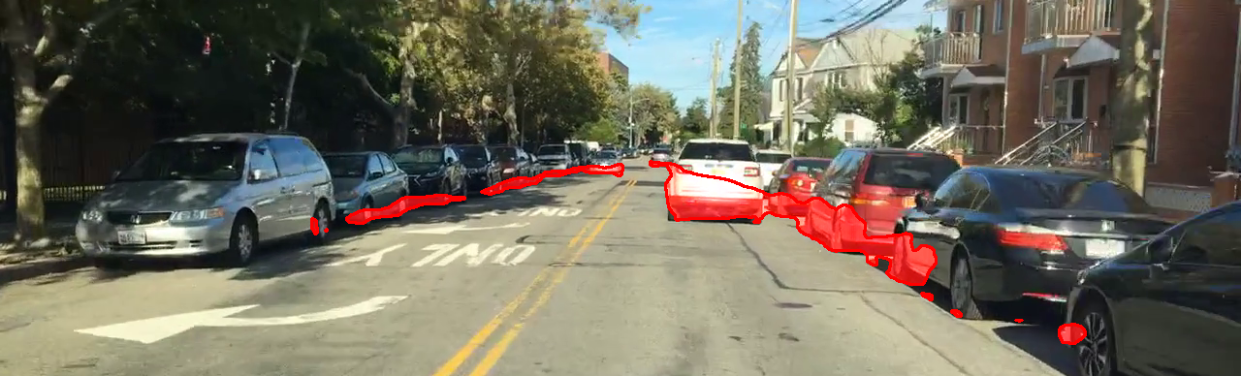}};
        \advance\vX by \vdX
        \node[inner sep=0pt] (a) at (\vX, \vY) {\INC{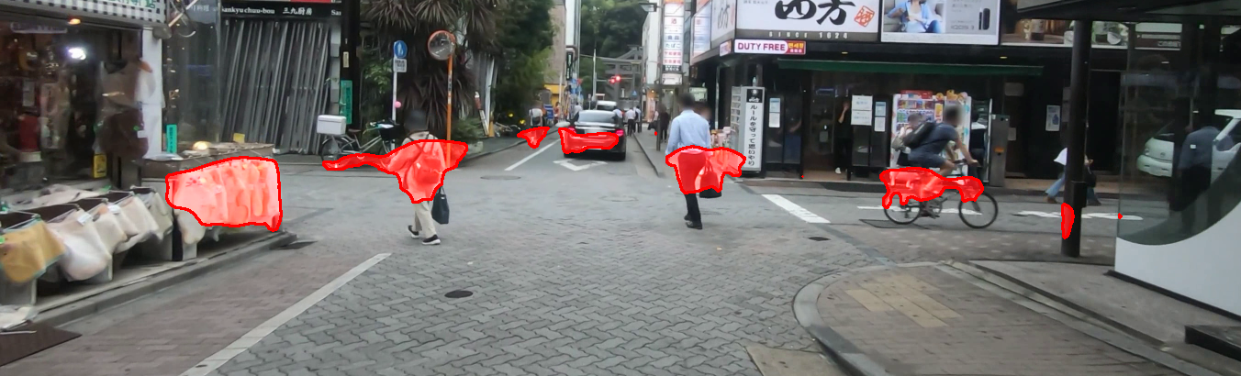}};

        \vX = 0
        \advance\vY by \vdY
        \node[inner sep=0pt, rotate=90,font=\tiny] (a) at (\vX,\vY) {\BSNshort\rule{0pt}{12pt}};
        \vX = 175
        \node[inner sep=0pt] (a) at (\vX, \vY) {\INC{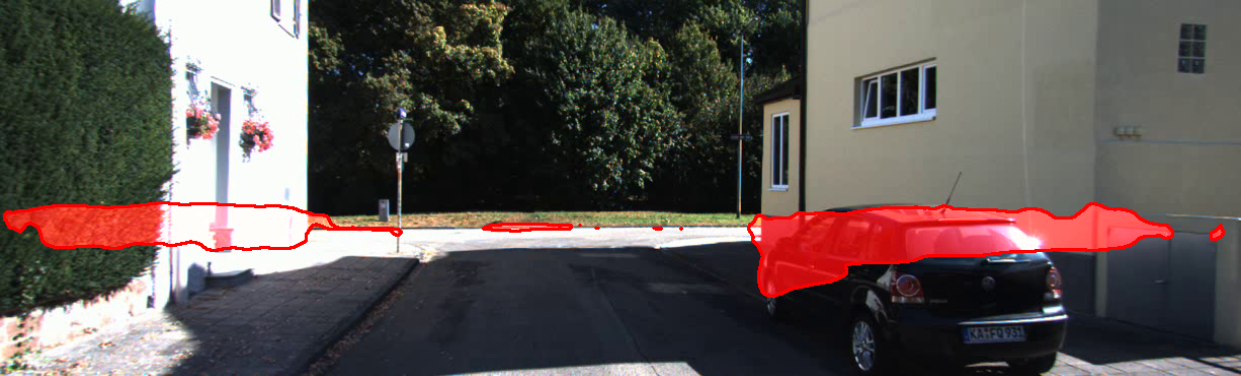}};
        \advance\vX by \vdX
        \node[inner sep=0pt] (a) at (\vX, \vY) {\INC{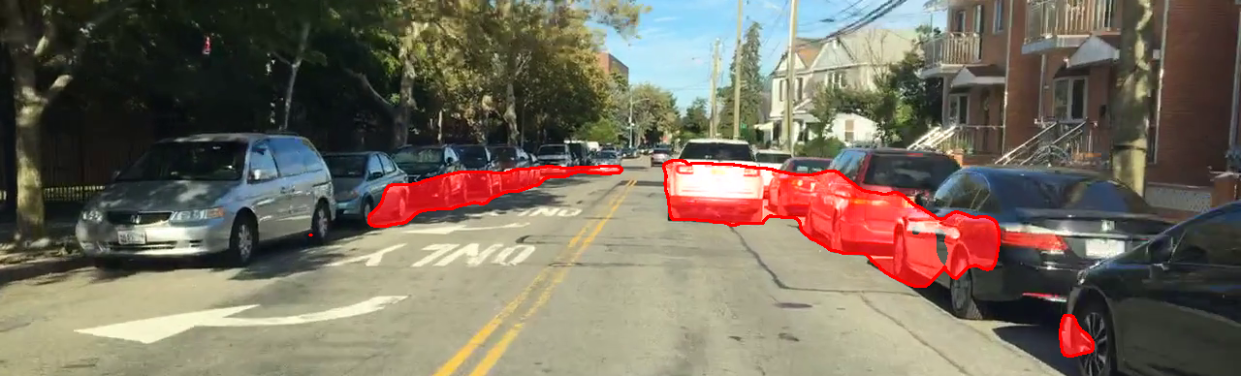}};
        \advance\vX by \vdX
        \node[inner sep=0pt] (a) at (\vX, \vY) {\INC{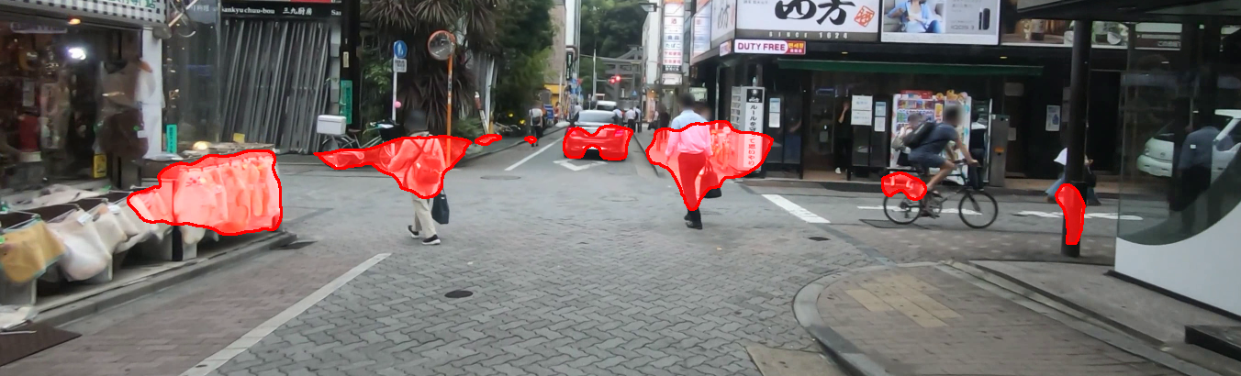}};

        \draw[thick] (-8,-368)--(964,-368)--(964,-682)--(-8,-682)-- cycle;
    \end{tikzpicture}
    \caption{Blind spot estimation results. From top to bottom, $T$-frame blind spots from our RBS Dataset, the results by \textit{Traversable}, \BSSS, \BSBOX, \BSNshort-D, \BSNshort-KD, and \BSNshort, respectively. Results are shown for  \BSNlong successfully achieves high precision and recall for complex road scenes (KITTI, BDD, and TITAN from left to right) by estimating nuanced blind spots caused by parked and moving cars, intersections, buildings, poles, gates, \etc. In the left most column, \BSNlong also estimates intersection blind spots that are even not in the ``ground-truth'' $T$-frame blind spots. This demonstrates the effectiveness of the visibility and the advantage of \BSNlong of being able to train on a diverse set of scenes thanks to the fact that $T$-frame blind spots can be easily computed on arbitrary driving videos.}
    \label{fig:qualitative}
\end{figure}
}

\subsubsection{Qualitative Evaluations}

\begin{figure}[t]
    \centering
    \begin{tabular}{c c}
        \includegraphics[width=0.49\linewidth]{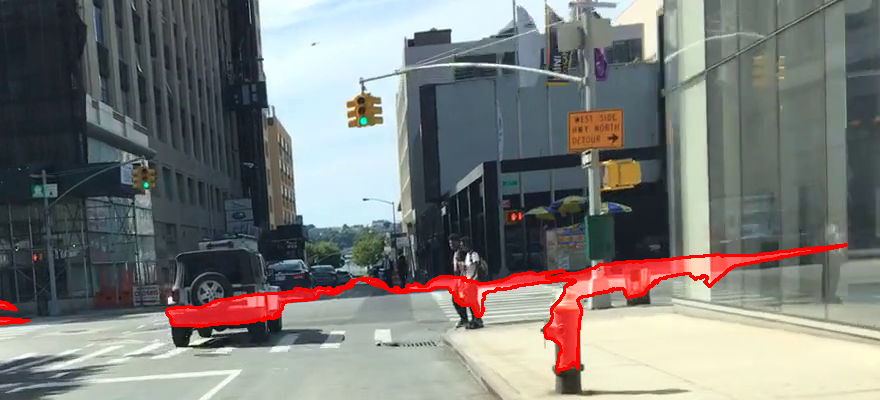}
        \includegraphics[width=0.49\linewidth]{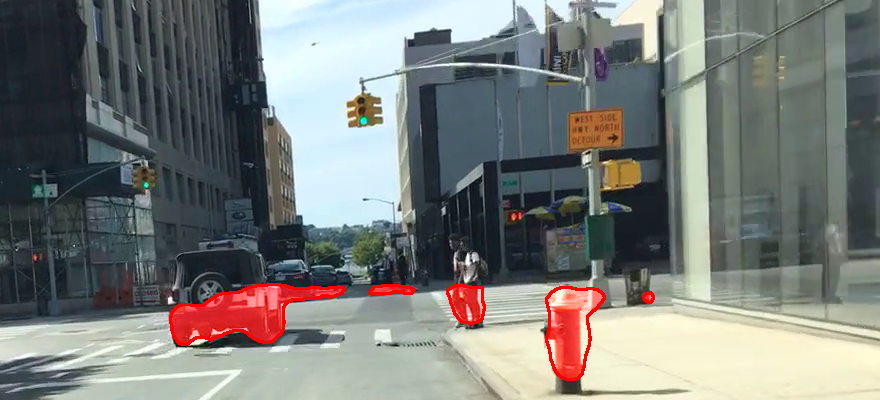} \\
    \end{tabular}
    \caption{Failure example. Left: $T$-frame blind spot (``ground truth'') from a frame in BDD100k-RBS. Right: Blind spot estimation results from \BSNlong trained on BDD100k-RBS. Due to the bias of BDD100k-RBS, which lacks left and right turns at intersections, \BSNshort cannot estimate the blind spots caused by the intersection. We plan to address these issues by employing a panoramic driving video for pre-training.}
    \label{fig:limitation}
\end{figure}
\cref{fig:qualitative} shows blind spot estimation results for the test sets. Compared with baseline methods, our method estimates the complex blind spots arising in these everyday road scenes with high accuracy. The ablation study results using \BSNshort-D and \BSNshort-KD clearly demonstrate the importance of the depth estimator and knowledge distillation in \BSNlong. It is worth noting that \BSNlong correctly estimates the left and right blind spots in the left column example, even though the ``ground-truth'' $T$-frame blind spots do not capture them due to the visibility map. These results clearly demonstrate the effectiveness of the visibility map and the training on diverse road scenes whose $T$-frame blind spots can be automatically computed. \BSNlong can be trained on arbitrary road scenes as long as $T$-frame blind spots can be computed, \ie, SLAM, semantic segmentation, and monocular depth estimation can be applied. In this sense, it is a self-supervised method.

\cref{fig:limitation} shows a failure case example. By definition of $T$-frame blind spots, \BSNlong cannot estimate intersection blind spots in videos that do not have any turns. We plan to explore the use of wider perspective videos, including full panoramic views, to mitigate this issue.

\subsubsection{Ablation Study}


\begin{table}[t]
    \centering
    \begingroup
    \setlength{\tabcolsep}{4pt}
    \tiny
    \caption{Ablation study.  Our model achieves comparable performance to larger models and its frame-rate is promising for realtime processing.}
    \label{tab:ablation}
    \begin{tabular}{c|cccccc}
    Architecture & IoU & Recall & Precision & \# of params & GMACS & FPS \\
    \hline
    U-Net based~\cite{ronneberger2015u}  & 0.289 & 0.484 & 0.417 & 17.3M & 280.2 & 139.9 \\
    Small (ours) & 0.330 & 0.563 & 0.444 & 18.1M &  47.1 & 37.5 \\
    Medium       & 0.315 & 0.478 & 0.482 & 32.0M &  93.0 & 19.5 \\
    Large        & 0.337 & 0.508 & 0.500 & 59.3M & 160.9 & 11.3 \\
    \end{tabular}
    \endgroup
\end{table}

We compare large/medium-sized blind spot estimators as well as a simple U-Net~\cite{ronneberger2015u} baseline with our small-sized (light-weight) blind spot estimator. The differences between the large/medium-sized blind spot estimators and the small-sized one are backbone, channel size, and the number of decoder layers. The backbones of the large/medium-sized ones are ResNet101 and ResNet50, respectively. 
\cref{tab:ablation} lists the IoU, recall, precision, the number of the parameters, the computational complexity in GMACS, and the inference speed with a single NVIDIA TITAN V 12GB.  The results show that our model achieves comparable performance to larger models with much smaller cost and runs sufficiently fast for real-time use.

\section{Conclusion}

We introduced a novel computer vision task for road scene understanding, namely 2D blind spot estimation. We tackle this challenging and critical problem for safe driving by introducing the first comprehensive dataset (RBS Dataset) and an end-to-end learnable network which we refer to as \BSNlong. By defining 2D blind spots as road regions that are invisible from the current viewpoint but become visible in the future, we showed that we can automatically compute them offline for arbitrary driving videos, which in turn enables learning to detect them with a simple neural network trained with knowledge distillation from a pre-trained semantic segmentation network. We believe these results offer a promising means for ensuring safer manual and autonomous driving and open new approaches to extending self-driving and ADAS with proactive visual perception.

%
%
\bibliographystyle{splncs04}
\bibliography{egbib}
\end{document}